\documentclass[letterpaper, 10 pt, conference, twocolumn, bottom=1in]{ieeeconf}
\usepackage[utf8]{inputenc}
\usepackage[english]{babel}
\usepackage{amsmath,empheq,amssymb}
\usepackage{mdframed}
\newmdtheoremenv{problem}{Problem}
\newcommand{\norm}[1]{\left\lVert#1\right\rVert}

\newcommand{\col}[0]{\mathrm{col}}

\usepackage{graphicx}
\usepackage[version=4]{mhchem}
\usepackage{siunitx}
\usepackage{longtable,tabularx}
\title{Smooth Time-Optimal Trajectories for Steering Drones}

\author{Srinath Tankasala$^{1}$, Can Pehlivanturk$^{1}$, Efstathios Bakolas$^{1}$, Mitch Pryor$^{1}$}

\usepackage[backend=bibtex,style=numeric,sorting=none]{biblatex}
\begin{document}

\maketitle

\begin{abstract}
    In this paper, we address a minimum-time steering problem for a drone modeled as point mass with bounded acceleration, across a set of desired waypoints in the presence of gravity. We first provide a method to solve for the minimum-time control input that will steer the point mass between two waypoints based on a continuous-time problem formulation which we address by using Pontryagin's Minimum Principle. Subsequently, we solve for the time-optimal trajectory across the given set of waypoints by discretizing in the time domain and formulating the minimum-time problem as a nonlinear program (NLP). The velocities at each waypoint obtained from solving the NLP in the discretized domain are then used as boundary conditions to extend our two-point solution across those multiple waypoints. We apply this planning methodology to execute a surveying task that minimizes the time taken to completely explore a target area or volume. Numerical simulations and theoretical analyses of this new planning methodology are presented. The results from our approach are also compared to traditional polynomial trajectories like minimum snap planning.
\end{abstract}

\section{Introduction}
The problem of steering a particle using an optimal controller with bounded inputs has received great attention in the literature, \cite{akulenko2007time, akulenko2002timesphere, venkatraman2006optimal}, for example. Several variants of the problem, such as steering a particle in the presence of a flow field have also been well-studied, \cite{bakolas2016time, bakolas2013optimal}. However, previous works don't address the case of continuous-time optimal steering of the particle in the presence of gravity. More importantly, most steering algorithms only consider the case of steering a particle between two points. There is no framework to extend the two-point solution to find a continuous-time optimal input for steering the particle across an arbitrary number of waypoints. In this work, we solve a minimum time optimal problem between two waypoints while taking gravity into account. We then provide a way to extend this solution to include multiple waypoints using a discrete-time problem formulation. By using this extension of the two point solution, we address the problem of steering a particle through a set of prescribed waypoints in minimum time. 
{\let\thefootnote\relax\footnote{{$^1$ Cockrell School of Engineering, The University of Texas at Austin\\
    301 E. Dean Keeton St. C2100
Austin, Texas 78712-2100}}}
% \footnotetext{\ $^1$ Cockrell School of Engineering, The University of Texas at Austin\\
%     301 E. Dean Keeton St. C2100
% Austin, Texas 78712-2100}
It is assumed that the magnitude of the control input of the particle is upper bounded by an \textit{a priori} fixed constant. Hence the thrust input value set for the particle is a \lq hypersphere\rq. The problem considered here belongs to a class of minimum-time problems for linear systems where the control input is constrained to a convex set \cite{lee1967foundations}. When the input set is constrained to a hypersphere, one can characterize the time optimal control input law. In our case, there always exists a time-optimal control input that attains its values in the boundary of the input value set. This is commonly known as the boundary control strategy, \cite{du2018boundary}, or bang-bang control in the special case when the input set is a hypercube. Several approaches for generating time-optimal trajectories for UAVs are available in the literature: Lai et al \cite{lai2006time} uses a discretization in the time domain to formulate the quadrotor steering problem as an non-linear program (NLP) that is solved using a genetic algorithm. The obtained solution is time optimal in the discretized input space, but that does not guarantee that as the time resolution of the discretization increases that one obtains the optimal solution to the continuous-time problem. 

In Loock et al, \cite{van2013time}, the trajectory is parameterized based on a set of reference geometric curves and a non-linear optimal control problem is formulated to calculate the minimum-time path. In \cite{traj:benchmark} the authors simplify the quadrotor dynamics to a 2D model and apply Pontryagin's Minimum Principle (PMP) to solve for the minimum-time solution. They assume a hypercube domain and use a bang-bang control strategy. The bang-bang strategy provides a solution where the system uses maximum actuator input along the boundary of the admissible control envelope. However, most actuators cannot instantaneously switch inputs and thus bang-bang solutions are difficult to track. Hence, a continuous time optimal input would make the generated trajectory smoother and easier to track.

The bang-bang control strategy in particular has found widespread application in the field of UAVs as a means to compute minimum-time trajectories \cite{traj:benchmark, lupashin2011adaptive}. There are many applications where steering the drone through a desired set of waypoints in minimum time is essential, such as inspection surveys, drone racing and package delivery, \cite{ZHOU202056, loianno2018special}. Quadrotors are however underactuated systems where some of the rotation and translation degrees of freedom are coupled. This means that maximum actuator input is not always used for accelerating the vehicle and some effort is spent on orienting the vehicle thrust (z-axis) correctly. Nonetheless, the minimum-time solution based on the point mass simplification of the drone dynamics serves as a starting point/low fidelity result in many planning strategies \cite{foehn2021time, ryou2020multi}. It allows us to evaluate how much of the theoretical speed potential does a certain control strategy utilize and identify maneuvers that can improve it. Hence the problem of time-optimal steering of a point mass with bounded thrust is now of great interest in the UAV community. To this end, we demonstrate our proposed trajectory generation strategy by applying it to solve a minimum-time surveying task. The trajectory generation method can be applied to other tasks as well such as drone racing, etc. 

\begin{figure}[ht]
    \centering
    \includegraphics[width=3in]{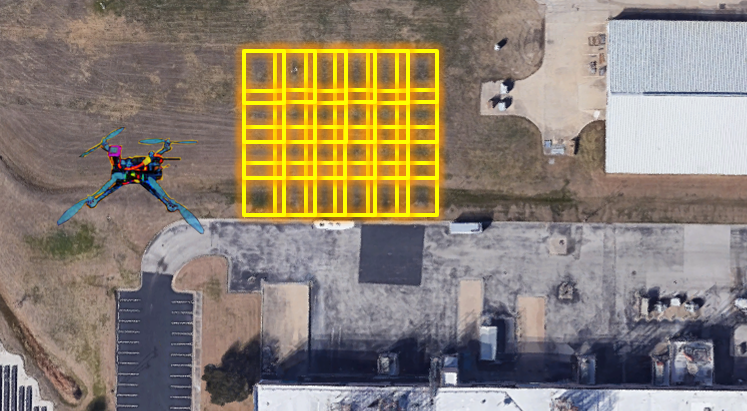}
    \caption{Representative pic of the drone and discretization of of survey area with defined image overlap (yellow rectangles) for stitching}
    \label{fig:overlapsurvey}
\end{figure}

%If the agent needs to be stationary at the start and end waypoints, the problem reduces to a boundary value problem (BVP) with given initial and terminal positions at zero velocity, and the optimal results are available in the literature \cite{traj:benchmark}. 
A common goal of a surveying task is to reconstruct a representation of the 3D scene (point cloud, textured map, etc.). To complete a survey of a given Region of Interest (RoI), the agent needs to pass through a known set of waypoints generated from a global planner. Here it is assumed that the drone does not need to halt at any of the intermediate waypoints. Determining the set of waypoints for complete coverage of the RoI depends on the information available about the terrain. Coverage path planning is well studied in the literature. Cabreira et al, \cite{coverage2019} presents a comprehensive study on the state of the art in coverage path planning methods. Standard coverage planning techniques apply cellular decomposition to divide the survey area into evenly sized cells (rectangles). Given the flight pattern, camera field-of-view (FoV), required ground resolution, and the desired image overlap, one can compute the cellular decomposition of the RoI (\lq A\rq) and corresponding waypoints that need to be visited for ensuring 100\% coverage of the desired area, as shown in Fig \ref{fig:overlapsurvey}.

The scope of the surveying task presented here is limited to offline planning, i.e. when information about the terrain layout (flat, hilly, urban, etc.) is available \textit{a priori}. The global planner is also responsible for calculating the optimal touring path that connects these waypoints, i.e. the order in which they will be visited. Depending on the user's objective (minimum time, energy, etc.) an appropriate touring path can be determined. An example of the cellular decomposition for our surveying task is shown in Fig \ref{fig:overlapsurvey}. The agent is required to pass through the center of the overlapping rectangles in a known order. % \subsection*{Contribution}

In this paper, we make three main contributions. Firstly, we present a systematic method to compute the time optimal trajectory for steering a drone of point mass from an initial position and velocity to a desired endpoint and final velocity in minimum time in the presence of gravity and bounded thrust. We do this by reducing the time-optimal control problem to a set of algebraic eqs that can be solved to obtain the minimum-time trajectory. Secondly, we provide a numerical method to address the minimum-time surveying problem by discretizing the problem in time, similar to \cite{lai2006time}. The discretized time domain formulation is framed as a non-linear program (NLP) that can be solved using standard optimization solvers. Thirdly, we lay out a framework to extend the solution from our synthesized time-optimal steering problem to a generalized waypoint navigation problem. We do this by using the waypoint velocities obtained from the NLP solution as a boundary condition for our trajectory generator. By applying this method, we achieve reduced survey times and smoother inputs making it easier to track on a real drone. The trajectories generated were faster than traditional planners, like minimum snap planning \cite{mellinger2011minimum}, completing the survey in $\approx$ 20\% lesser time.\\

In Section \ref{sec:implementation}, we present the dynamic model and solution to the time-optimal steering problem in the presence of gravity. In Section \ref{sec:surveying traj generation}, we discuss the minimum-time surveying problem as an NLP and extend our time-optimal solution to multiple waypoints. In Section \ref{sec:sim_results}, we perform simulations to demonstrate the efficiency of the trajectory by comparing the NLP solution and the minimum time solution obtained by PMP. We also show comparisons with traditional minimum snap solutions. Section \ref{sec:conclusion} summarizes our conclusions and discusses avenues for future extension in the planning algorithm.
\section{Problem Formulation}
\label{sec:implementation}
\subsection{Analysis of the optimal steering problem under gravity}
We consider the dynamics of the drone as a double-integral plant traveling in 3D space in a constant gravitational field, \textbf{g}. The magnitude of control input (acceleration) for steering the drone is upper bounded, $\bar{u}$. We assume the control input has no lower bound, i.e. zero control input is allowed. To simplify, we take the initial time, $t_0 = 0,\ r(0) = r_0$, and $v(0)=v_0$. Then we can assume the motion of the drone is described by the following set of eqs:\\
\begin{equation}
\begin{split}
    &\dot{{r}}(t) = {v(t)}, \quad {\dot{v}(t)} = {u(t) + g}, \\
    &{\norm{u(t)}} \leq \bar{{u}}\\
\end{split}\label{eq:sys}
\end{equation}
where ${r(t)}\in \mathcal{R}^3 ({r_0} \in \mathcal{R}^3)$ and ${v(t)}\in \mathcal{R}^3 ({v}_0 \in \mathcal{R}^3)$ are the position and velocities of the drone respectively at time ${t}$ (time ${t=0}$), ${u(t)}$ is the control input at time ${t}$ and $\bar{{u}}$ and ${g}$ are non-negative constants. \\
\textbf{Definition 1} 
We refer to the composite state of the drone as $z$, which is a concatenation of the position and velocity vectors, ${z(t)}:=\col({r(t)},{v(t)})$. The initial state of the drone is given by ${z(0)=z_0}=\col({r_0},{v_0})$.

We rewrite the system in \eqref{eq:sys} according to the convention in Athans \& Falb, \cite{athans2013optimal} as follows:\\
\begin{equation}
\begin{split}
    % &\begin{bmatrix} \dot{{r}}(t) \\ \dot{{v}}(t) \end{bmatrix}= {f(v}(t))+{Bu(}t)\\
    &{\dot{z}(t)}= {f(z)+Bu(t)}\\
    &{\norm{u(t)} \leq \bar{u}}\\
    \text{where, } &{f(z)}=\begin{bmatrix} {v} \\ {g} \end{bmatrix}; {B = \begin{bmatrix} {0_{3\times 3}} \\ {I_{3\times 3}} \end{bmatrix}}; \\
    &\text{and, }{z(0) = z_0}
\end{split}\label{eq:generic sys}
\end{equation}
We wish to steer the drone from $z_0$ at time $t=0$ to the final state ${z_f = (r(t_f),v(t_f)) = (r_f,v_f)}$ at time ${t=t_f}$ for a given input ${u(t)}$.\\
\textbf{Definition 2} Given a initial state $z_0$ and final state $z_f$ of the drone, the set of all admissible control input functions that steers the drone from start is given by $\mathcal{U}$. In other words:\\
% $\mathcal{U}=\{{u(t)}\ {s.t.}\ ||{u(t)}||\leq \bar{{u}}\ \&\ {u(t):z_0\rightarrow z_f}\}$\\
$\mathcal{U}$ is the set of all piecewise continuous functions of time, $u(t)$, s.t. $||{u(t)}||\leq \bar{{u}}\ \forall\ t\in [0,t_f]$ and ${u(t):z_0\rightarrow z_f}$.

For the system given in eq \eqref{eq:generic sys}, the Hamiltonian for the time optimal steering problem can be written as:\\
\begin{align}
    \mathcal{H}{(t,z,u,\lambda) = 1 + \lambda^T(f(v)+Bu(t))}
\end{align}
where $\mathcal{H}{(\cdot):[0,\infty)}\times \mathcal{R}^6\times \mathcal{R}^3 \times \mathcal{R}^6\mapsto \mathcal{R}$, and  ${\lambda: [0,t_f]} \mapsto \mathcal{R}^6$ is the co-state vector. For simplification purposes, we rewrite ${\lambda(\cdot)}$ as:
\begin{align*}
    {\lambda = \col(p,q)};\ \ \text{where}\ \ {p,q: [0,t_f]} \mapsto \mathcal{R}^3
\end{align*}
Thus the Hamiltonian for the system in eq \eqref{eq:generic sys} can be rewritten as:
\begin{align}
    \mathcal{H} = {1 + p^T v + q^T (u + g)}.
\end{align}

Next, we apply Pontryagin's minimum principle (PMP), \cite{athans2013optimal}, to characterize the time-optimal steering input. Let ${z^*(\cdot):[0,t_f^*]} \mapsto \mathcal{R}^6$, where ${z^*(\cdot)}=\col({r^*(\cdot),v^*(\cdot))}$ denote the minimum-time trajectory generated by applying the time optimal control input ${u^*(\cdot)}$ over the interval ${[0,t_f^*]}$. Then, the existence of a continuous co-state function ${\lambda^*(\cdot):[0,t_f^*]}\mapsto \mathcal{R}^6$ corresponding to the optimal trajectory ${z^*(\cdot)}$ is guaranteed and it neccessarily satisfies the following system of differential eqs:
\begin{align}
    \begin{split}
    &{\dot{\lambda}^*} = -\left(\nabla_{z} \mathcal{H} \right)^T = \begin{bmatrix}-\left(\frac{{\partial} \mathcal{H}}{{\partial x}}\right)^T\\
    \\
    -\left(\frac{{\partial} \mathcal{H}}{{\partial v}}\right)^T\end{bmatrix} \\
    \end{split}\label{eq:optimal_lambda}
\end{align}
    Let ${\lambda^*} = \col{(p^*,q^*)}$, then from eq \eqref{eq:optimal_lambda}, it follows that:
\begin{align}
    &\begin{bmatrix}{\dot{p}^*}\\{\dot{q}^*}\end{bmatrix} = \begin{bmatrix}0\\ {-p^*}\end{bmatrix}\label{eq:costate dot},\\
    \therefore\ &{\lambda^*(t)} = \begin{bmatrix}{p_c}\\ {-p_ct+q_c}\end{bmatrix},\ \  {\forall\ t\in[0,t_f^*]}\label{eq:costate}
\end{align}
where $p_c, q_c \in \mathcal{R}^3$ are constants. The time-optimal control $u^*(\cdot)$ necessarily minimizes the Hamiltonian evaluated along the optimal state ($z^*(\cdot)$) and costate ($\lambda^*(\cdot)$) trajectories, that is,
\begin{equation}
    {\langle u^*(t), B^T \lambda^*(t) \rangle \leq  \langle v, B^T \lambda^*(t) \rangle\ \forall\ \{v\in\mathcal{R}^3\ \mathrm{s.t.}\ \norm{v}\leq \bar{u}\} } \label{eq:pmp_ineq}
\end{equation}
where ${u^*(t)}$ is the optimal control input, and ${\lambda^*(t)}$ is the solution of eq \eqref{eq:costate dot} that corresponds to the optimal trajectory ${z^*(t)}$ and optimal input ${u^*(t)}$. Using eq \eqref{eq:pmp_ineq}, the optimal input is given by,
\begin{align}
    & {u^*(t)} = \begin{cases}
    -\frac{{B^T\lambda^*(t)}}{\norm{{B^T\lambda^*(t)}}} \ \quad \text{if}\ {B^T\lambda^*(t)\neq 0}\\
    {v} \in \mathcal{U}\quad \quad \quad \quad \text{otherwise}
    \end{cases}
\end{align}
In view of eq \eqref{eq:costate}, the optimal input can be written as:
\begin{align}
    & {u^*(t)} = \begin{cases}
    {\bar{u}\frac{-p_c t+q_c}{\norm{-p_ct+q_c}}} \ \quad \text{if}\ {-p_ct+q_c\neq 0}\\
    {v} \in \mathcal{U}\quad \quad \quad \quad \text{otherwise}
    \end{cases}\label{eq:optimal input}
\end{align}
For the complete derivation of eq \eqref{eq:optimal input} please refer to \cite{athans2013optimal}.

\subsection{Navigation formula for time-optimally steering a point mass to a specified position and velocity}
Given the form of the minimum-time control input in eq \eqref{eq:optimal input} which was derived using PMP, the corresponding time-optimal trajectory can be found for a given set of boundary conditions. For convenience, the eqs in (\ref{eq:sys}) can be non-dimensionalized w.r.t. ${\bar{u}}$, i.e. $\mathrm{v}= v/\bar{u}$ and $\mathrm{r}=r/\bar{u}$. Thus the non-dimensionalized set of eqs becomes:
\begin{align}
\begin{split}
    &\mathrm{\dot{r}(t) = v(t)}; \quad \mathrm{\dot{v}(t) = u(t) + g}\\
    &\mathrm{\norm{u(t)} \leq 1}
\end{split}\label{eq:nondim sys}
\end{align}
where, $\mathrm{g}=g/\bar{u}$ is the non-dimensionalized gravity.\\
The form of $u^*(t)$ is unaffected by gravity acceleration as shown in eq \eqref{eq:optimal input}. However, analytically solving for the optimal input $u^*(t)$ that satisfies a given boundary condition, $\col(r_0,v_0),\ \col(r_f,v_f)$ is non-trivial because of its form. This is why most minimum-time solutions adopt alternative methods (numerical), \cite{lai2006time, traj:benchmark}. The analytical solution for the case of time optimal steering without gravity has been solved in Akulenko et al, \cite{akulenko2007time}. Here these eqs are extended to account for gravity. For completeness, all eqs with the gravity terms below use the same notation as \cite{akulenko2007time}.
We define the following parameters in order to solve the optimal input for the system in eq \eqref{eq:nondim sys}:
\begin{align}
    &u := \frac{-p_0t+q_0}{\norm{-p_0t+q_0}}\label{eq:optim_prb_defs}\notag\\
    &\xi := \frac{p_0}{\norm{q_0}};\quad \eta:=\frac{q_0}{\norm{q_0}}\notag\\
    &\rho := \norm{\xi}\neq 0;\quad \sigma := \frac{\xi \cdot \eta}{\norm{\xi}\norm{\eta}} = \frac{\xi \cdot \eta}{\norm{\xi}}\ .
\end{align}
where $p_0$, $q_0\in \mathcal{R}^3$ are constant vectors. Using the definitions in eqs \eqref{eq:optim_prb_defs}, the optimal input from eq \eqref{eq:optimal input} is written as:
\begin{flalign}
    &\quad \quad \quad \quad \ u(t) = \frac{Q(t)}{R(t)},&&\\
    &\text{where  }\ Q(t) := -\xi t+\eta;\quad R(t) := \sqrt{\rho^2 t^2 - 2\sigma \rho t + 1}.&&\notag%\label{eq:int ut} 
    % &\dot{v} = u + g;&&\label{eq:vt}
\end{flalign}
Integrating the RHS of eqs in \eqref{eq:nondim sys} results in: %\eqref{eq:int ut} and \eqref{eq:vt}:
\begin{align}
    &v(t) = v_0 + \int_0^{t}\frac{Q(\tau)}{R(\tau)}d\tau + gt \label{eq:int vt},\\
    &r(t) = r_0 + v_0t + \frac{1}{2}gt^2+ \int_0^{t}\frac{(t-\tau)Q(\tau)}{R(\tau)}d\tau \label{eq:int xt} .
\end{align}
For any instance of time $t$, the drone's velocity and position in presence of gravity can be written as:
\begin{equation}
    \begin{split}
        &v(t) = v_0 + gt + V_{\xi}(t)\xi + V_{\eta}(t)\eta,\\
        &r(t) = r_0 + v_0t + \frac{1}{2}gt^2 + X_{\xi}(t)\xi + X_{\eta}(t)\eta,\\
        \end{split}\label{eq:solution form}
\end{equation}        
where
\begin{equation}
    \begin{split}
        &S(t) := \log\left(k + \sqrt{1+k^2}\right);\ k = \frac{(\rho t-\sigma)}{\sqrt{1-\sigma^2}},\\
        &V_{\xi}(t) := -\frac{1}{\rho^2}\left[\sigma S(\tau) + R(\tau)\right]_0^t; V_{\eta}(t) = \frac{1}{\rho}S(\tau)|_0^t,\\
        &X_{\eta}(t) := \frac{1}{\rho^2}\left[-R(\tau)+(\rho\tau-\sigma)S(\tau)\right]_0^t-\frac{t}{\rho}S(0),\\
        &X_{\xi}(t) := 
        \frac{1}{2\rho^3}\left[(\rho\tau+3\sigma)R(\tau)\right]_0^t+\frac{t}{\rho^2}(1+\sigma S(0))\\
        &\quad \quad \quad \quad +\left[(-2\rho\sigma\tau+3\sigma^2-1)S(\tau)\right]_0^t.\\
        % &\quad \quad \quad \quad +\frac{t}{\rho^2}(1+\sigma S(0)).     
    \end{split}\label{eq:optim integrals}
\end{equation}
The trajectory must start at time $t=0$ from $r_0$ with velocity $v_0$ in the non-dimensional space and end at $r_f$ with non-dimensional velocity $v_f$. Assuming the time taken to complete this path is given by $t_f$, these boundary conditions can be substituted to obtain the eqs to be solved, i.e.:
\begin{flalign}
    &-r_f+r_0+v_ft_f-g\frac{t_f^2}{2} = t_f^2(a_\zeta \zeta + a_\eta \eta),\label{eq:optim disp eqs}&&\\
    &v_0-v_f+gt_f = t_f(b_\zeta \zeta + b_\eta \eta),\label{eq:optim vec eqs}&&\\
    &\text{where  } \notag\\
    &\zeta := \xi t_f,\quad \norm{\zeta} := \mu = \rho t_f,\notag\\
    &a_\zeta = -\frac{(\mu+3\sigma)a+\mu+(3\sigma^2-1)b}{2\mu^3},&&\notag\\ 
    &a_\eta = b_\zeta = \frac{a+\sigma b}{\mu^2};\ b_\eta = -\frac{b}{\mu},\ a = \sqrt{\mu^2-2\mu\sigma+1}-1, &&\notag\\
    &b = \log\left(\frac{\mu-\sigma+\sqrt{\mu^2-2\mu\sigma+1}}{1-\sigma}\right)\ .&&\notag
\end{flalign}
In order to solve eqs \eqref{eq:optim disp eqs} and \eqref{eq:optim vec eqs}, we convert them into 3 scalar eqs by taking dot product of the vectors with respect to each other, i.e. LHS(\eqref{eq:optim disp eqs})$ \cdot $LHS(\eqref{eq:optim disp eqs}) $=$ RHS(\eqref{eq:optim disp eqs}) $\cdot$ RHS(\eqref{eq:optim disp eqs}). Similar dot products are applied between (\eqref{eq:optim vec eqs},\eqref{eq:optim vec eqs}) and (\eqref{eq:optim disp eqs},\eqref{eq:optim vec eqs}) giving:
\begin{align}
    & t_f^2f_{vv} = \norm{v_0-v_f}^2 + \norm{g}^2 t_f^2 + 2(v_0-v_f)\cdot gt_f\label{eq:fvv}\\
    \begin{split}
    &  t_f^4f_{xx} = \norm{r_0}^2 + \norm{v_f}^2t_f^2 + \frac{\norm{g}^2}{4}t_f^4 \\
    & \quad \quad \quad - g\cdot v_f t_f^3 - g\cdot r_0 t_f^2 + 2x_0\cdot v_f t_f \\
    \end{split}\label{eq:fxx}\\
    \begin{split}
    &f_{xv}t_f^3 = r_0\cdot (v_0-v_f) + (r_0\cdot g) t_f\\
    & \quad \quad  \quad + (v_f \cdot (v_0-v_f)) t_f + (v_f\cdot g)t_f^2\\
    & \quad \quad  \quad - g\cdot (v_0-v_f) \frac{t_f^2}{2} - \norm{g}^2\frac{t_f^3}{2}\\
    \end{split} \label{eq:fxv}
\end{align}
By simplifying eqs \eqref{eq:fvv}-\eqref{eq:fxv}, we obtain the following set of 3 scalar eqs that need to be solved in terms of $\eta, \xi, t_f$ that determine the time-optimal trajectory:
\begin{flalign}
\begin{split}
    &r_0\cdot (v_0-v_f) = -(r_0\cdot g) t_f - (v_f \cdot (v_0-v_f)) t_f \\
    &\quad \quad \quad \quad \quad + g\cdot (v_0-3v_f) \frac{t_f^2}{2}\\
    &\quad \quad \quad \quad \quad + \left(f_{xv}+\frac{\norm{g}^2}{2}\right)t_f^3 \ \label{eq:vel_disp}\\
\end{split}\\
    &\norm{v_0-v_f}^2 = t_f^2(f_{vv} - \norm{g}^2)-2((v_0-v_f)\cdot g)t_f\  \label{eq:vel}\\
    \begin{split}
    &\norm{r_0}^2 = t_f^4(f_{xx} - \frac{\norm{g}^2}{4}) + (g\cdot v_f) t_f^3 \\
    & \quad \quad \quad - \left(\norm{v_f}^2 - g\cdot r_0\right)t_f^2 - 2(r_0\cdot v_f) t_f\label{eq:disp}
    \end{split}
\end{flalign}
where 
\begin{flalign*}
    &f_{vv} = (b^2_\zeta \mu^2 + b_{\eta}^2 \sigma^2 + 2b_\eta b_\zeta \mu \sigma),\\
    &f_{xx} = (a^2_\zeta \mu^2 + a_{\eta}^2 \sigma^2 + 2a_\eta a_\zeta \mu \sigma),\\
    &f_{xv} = a_\zeta b_\zeta \mu^2 + (a_\zeta b_\eta + a_\eta b_\zeta)\mu\sigma + a_\eta b_\eta\ .
\end{flalign*}
Eqs \eqref{eq:vel_disp}-\eqref{eq:disp} can be solved for the three scalar unknowns $\mu$, $\sigma$ and $t_f$. The bounds of the variables are $\mu>0$, $-1<\sigma<1$ and $t_f>0$. These three parameters can then be used to calculate $\eta, \xi$, from eqs \eqref{eq:optim disp eqs} and \eqref{eq:optim vec eqs} which gives the full trajectory between two waypoints by substituting them in eqs \eqref{eq:solution form}%\eqref{eq:int vt} and \eqref{eq:int xt}.

\section{Trajectory generation for surveying task}
\label{sec:surveying traj generation}

For the surveying task, we assume that the touring path is specified by the user (zig-zag, spiral, circular, etc.). The zig-zag survey flight pattern is used for our simulations and analysis. We consider coverage paths composed of a sequence of waypoints and a prescribed upper bound on the drone speed at the given waypoints. This is done to avoid motion blur in the images captured by the drone when passing through each waypoint. The objective is to generate the trajectory that utilizes maximum thrust from the quadrotor to minimize total survey time. For this, we use our trajectory generator as the local planner responsible for computing the intermediate keyframes (position, velocity, acceleration and yaw). 

\textbf{Definition 3} For a RoI \lq$A$\rq, the optimal touring path for the survey is given by $W(A) = {w_1, w_2,..., w_n}$ where, $w_k \in \mathcal{R}^{3}\ \forall k$, where $n$ is the total number of waypoints.

The time-optimal coverage trajectory generation problem can be formulated as:
\begin{problem}\label{prb:general} 
Time optimal coverage trajectory generation for a region of interest 'A' is
% \[
% \begin{gathered}
% \centering
  \begin{align}
  &{\min_{\Omega\in \mathcal{R}^{4}_+} \int_{0}^{t_N} dt}  \notag\\ 
  {s.t.:}\ \ &{\dot{x}(t) = f(x,\Omega)} \label{eq: system dyn}\\
  &{r(t_i) = w_i ,\quad w_i \in} W(A) \label{eq: waypoint constraint}
  %\quad &{\norm{v(t_i)} \leq \bar{v}_{blur}} \label{eq: speed constraint}
  \end{align}
% \end{gathered}
% \]
\end{problem}

 \noindent where ${x(t) = \col(r(t),v(t),q(t),\omega(t))} \in \mathcal{R}^{14}$ is the state (column) vector, consisting of the position ${r(t)}\in \mathcal{R}^3$, velocity ${v(t)}\in \mathcal{R}^3$, the orientation represented by the quaternion ${q(t)\in}  \mathcal{R}^4$, and the angular velocity ${\omega(t)\in}  \mathcal{R}^3$. The input in such a system is the vector ${\Omega(t)\in \mathcal{R}^4_+}$ where ${\Omega_j (t)}$ is the speed of the ${j^{th}}$ rotor. The non-linear dynamics are expressed in \eqref{eq: system dyn}, where $f(x,\Omega)$ is the translational and rotational dynamics of a quadrotor described in \cite{mellinger2011minimum}.
 
 The constraint \eqref{eq: waypoint constraint} corresponds to the imposed waypoint positions. Since the quadrotor dynamics are differentially flat, \cite{mellinger2011minimum}, i.e. the states and the inputs of the original system can be written as algebraic functions of flat outputs and their derivatives, position ${r} \in \mathcal{R}^3$, and yaw angle ${\psi} $, we plan the trajectory in the flattened output space. We assume yaw angle, ${\psi}$, is always 0. The limitations of the original system are usually addressed by enforcing additional constraints such as maximum jerk, to account for the thrust, angular velocity and angular acceleration limits of the quadrotor. The optimization method presented here can be extended to include such constraints. In this work, we consider only the particle dynamics and consider a state vector, ${x(t) = \col(r(t),v(t))}$. Hence the system dynamics ${f(x,\Omega)}$ simplifies to the system of eqs shown in \eqref{eq:generic sys}.

\subsection{Waypoint Velocity Determination}
The first step in computing the time-optimal trajectory in Problem \ref{prb:general} is to find the optimal velocities the drone must have at each desired waypoint. We then interpolate the optimal waypoint velocities to get a smooth trajectory. To determine the optimal waypoint velocities, we discretize the domains of Problem \ref{prb:general}. To approximate the optimal waypoint velocities, we assume the input between two waypoints is of the form:

\begin{equation}\label{eq:constinput}
 {u(t) = }
  \begin{cases} 
   {\bar{u} \frac{\eta_{n,1}}{\norm{\eta_{n,1}}}} &  {t_n \leq t \leq t_{n,s}} \\
   {\bar{u} \frac{\eta_{n,2}}{\norm{\eta_{n,2}}}} &  {t_{n,s} < t < t_{n+1}}
  \end{cases},
\end{equation}
where ${n \in \{1,2,...,N\}}$ is the waypoint index, ${t_n}$ denotes the time at the ${n^{th}}$ waypoint and ${t_{n,s}}$ is the switching time where the input switches from one constant vector to another. The control input can be represented in a discrete form as follows:
\begin{flalign}\label{eq:discreteinput}
 {u_k = \left[\bar{u} \frac{\eta_{1,1}}{\norm{\eta_{1,1}}} ,  \bar{u} \frac{\eta_{1,2}}{\norm{\eta_{1,2}}}, ... , \bar{u} \frac{\eta_{N-1,1}}{\norm{\eta_{N-1,1}}}, \bar{u} \frac{\eta_{N-1,2}}{\norm{\eta_{N-1,2}}}\right]^T }&&
\end{flalign}
with variable discrete time steps:
\begin{flalign}\label{eq:discretetimestep}
 {dt_k =\left[t_{1,s}-t_1,t_{2}-t_{1,s},...,t_{N-1,s}-t_{N-1}, t_N-t_{N-1,s} \right]}&&
\end{flalign}
where ${k \in \{1,2,...,2(N-1)\}}$. The velocities and positions at respective time steps can be determined through integration:
\begin{align}
    &{r_{k+1} = r_k + v_k dt_k + u_k\frac{dt_k^2}{2}} \label{eq:discreteintegral}\\
    &{v_{k+1} = v_k + u_k dt_k} \notag
\end{align}
Using discretized inputs and flight times, the time-optimal trajectory of Problem \ref{prb:general} is cast as the NLP in Problem \ref{prb:nlp}. 
\begin{problem}\label{prb:nlp} 
Discretized time-optimal trajectory NLP
\[
\begin{gathered}
  \begin{aligned}
  &{\min \sum_{k = 1}^{2N -1} dt_k}\\
  {s.t.}\ \ \ &{r_{k+1} = r_k + v_k dt_k + u_k\frac{dt_k^2}{2}}\\
  &{v_{k+1} = v_k + u_k dt_k}\\
  &{r_j = w_i ,\ \ i  = 1,...,N}\\
  &\quad \quad \quad \quad \ {j = 1, 3, ..., 2N-1}\\
  &{\norm{u_k} = \bar{u}} \\
  \end{aligned}
\end{gathered}
\]
\end{problem}
The method to solve such a switching point NLP is known in the literature. The NLP problem here can be solved using an Interior Point Algorithm for Nonlinear Optimization (IPOPT) \cite{IPOPT} to find the required waypoint velocities. 

\subsection {Optimal Trajectory Generation}
After calculating the optimal velocities at each waypoint by solving Problem \ref{prb:nlp}, we will use them to generate the continuous-time optimal input for the drone. The continuous-time optimal input will generate the trajectory (position, velocity, acceleration) passing through the prescribed waypoint positions and with the calculated optimal waypoint velocities. The common approach to generate the continuous trajectory is to directly integrate the solution to Problem \ref{prb:nlp} between the time steps. 

\begin{equation}\label{eq:interpolate}
\begin{aligned}
 {u(t)} &= {u_k};\quad {v(t)} =  {v_k +  u_k t}\\  
 {r(t)} &= {r_k + v_k t +  u_k \frac{t^2}{2}}\\
 {k} &=
 \begin{cases}
 {0} & {0 \leq t \leq \sum_{k=1}^{1} dt_k}\\
 {1} & {\sum_{k=1}^{1} dt_k \leq t \leq \sum_{k=1}^{2} dt_k}\\
 {\vdots} & {\vdots}\\
 {2N-2} & {\sum_{k=1}^{2N-2} dt_k\leq t \leq \sum_{k=1}^{2N-1} dt_k}
 \end{cases}
\end{aligned}
\end{equation}

While this approach is easier to solve, it does not give us any indication of how far the solution is from the true continuous-time optimal solution. To obtain the true time-optimal solution between waypoints, one needs to solve the optimal control problem by utilizing PMP and computing the input that minimizes the Hamiltonian at each instant along with the optimal state and co-state trajectories. This can be done by using the NLP generated velocities at each waypoint ($v_0, v_2, v_4,... v_{2N-2}$). We use these velocities as boundary conditions for the time-optimal steering problem whose solution was presented in Section \ref{sec:implementation}

\begin{equation}\label{eq:interpolate cont}
\begin{aligned}
 {u^*(t)} &= {\bar{u}\frac{-p_it+q_i}{\norm{-p_it+q_i}}}  \\
 {v(t)} &=  {v_i +  \int_0^t u^*(t) dt}\\
 {r(t)} &= {r_i + \int_0^t v(t) dt}\\
 {i} &=
 \begin{cases}
 {0} & {0 \leq t \leq \sum_{k=0}^{1} dt_k}\\
 {1} & {\sum_{k=0}^{1} dt_k \leq t \leq \sum_{k=0}^{2} dt_k}\\
 \vdots & \vdots\\
 {N-1} & {\sum_{k=0}^{N-1} dt_k\leq t \leq \sum_{i=0}^{N} dt_k}
 \end{cases}
\end{aligned}
\end{equation}
We solve the system of eqs in \eqref{eq:interpolate cont} in terms of the unknowns ${dt_i, p_i}$ and ${q_i}$ for each set of waypoints. The optimal inputs obtained from $p_i$, $q_i$'s steers the drone between two waypoints (${r_{i-1}}$ to ${r_i}$) with the start(${v_{i-1}}$) and end velocities(${v_i}$). The ${v_i}$'s come from solving the NLP in Problem \ref{prb:nlp} and ${r_i}$'s come from waypoint locations ${w_i}$.

\section{Simulation results}
\label{sec:sim_results}

For our simulations, we assume the drone mass to be $1\ kg$, the maximum magnitude of the thrust of the drone to be $10.5 N$ and acceleration due to gravity as $g = 9.8 m/s^2$. The set of optimal inputs and times ($dt_i$'s) that solve eqs \eqref{eq:interpolate cont} were computed using the trust region dogleg algorithm in MATLAB r2021b. Depending on the boundary conditions, the problem may be ill conditioned for trust region based optimizers and multiple starting points are needed for the optimizer. The search space was narrowed by using upper bounds on time ($dt_i$'s) and bounds on $\mu$ and $\sigma$. Multiple solutions existed even at residual tolerance values of order $10^{-6}$. To obtain a a sufficiently accurate result, the residual was reduced to an order of $10^{-11}$.
% After obtaining the 3 unknowns, they are substituted back into equation \ref{eq:optim vec eqs} to obtain the vectors $\zeta$ and $\eta$ and hence $u^*(t)$.
\subsection{Comparison of interpolation methods}

It is of interest to see how good is the solution obtained from discretizing the search domain when solving for the optimal trajectory. The NLP solution of Problem \ref{prb:nlp} produces the required waypoint velocities. To interpolate with the optimal trajectory, we use those velocities as the boundary conditions for the optimal control problem solved earlier using PMP(refer to Section \ref{sec:implementation}). The entire domain was non-dimensionalized w.r.t. $\bar{u}$ and the minimum-time solution based on PMP is computed. The resultant optimal positions, velocities and accelerations are then scaled back to compare how good a direct interpolation is. 

\begin{figure}[!h]
    \centering
    \includegraphics[width=3in]{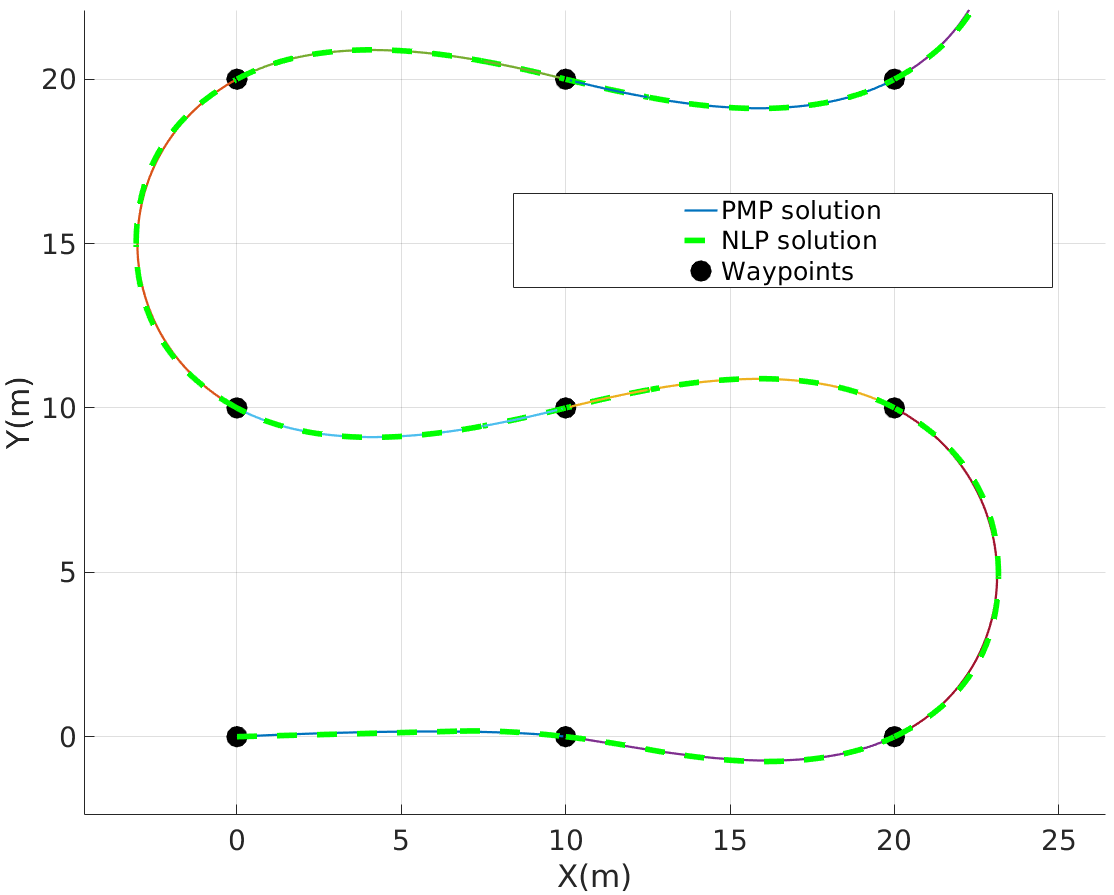}
    \caption{Paths obtained from direct interpolation vs continuous optimal input solution (paths truncated for clarity)}
    \label{fig:traj_comp}
\end{figure}

\begin{figure}[!h]
    \centering
    \includegraphics[width=3.1in]{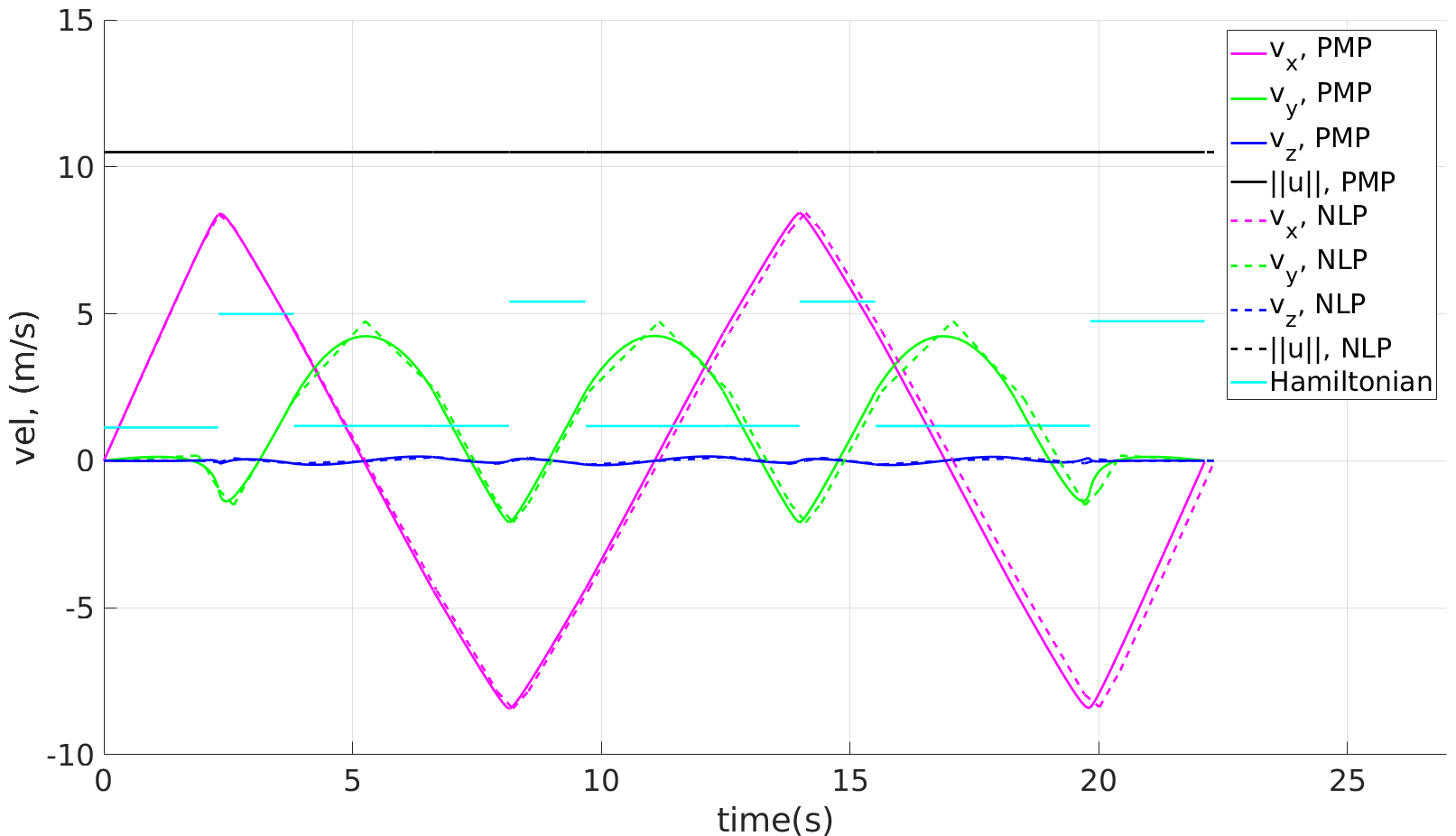}
    \caption{Velocity components from direct interpolation and continuous optimal input solution}
    \label{fig:vel_comp}
\end{figure}

\begin{figure}[!h]
    \centering
    \includegraphics[width=3.1in]{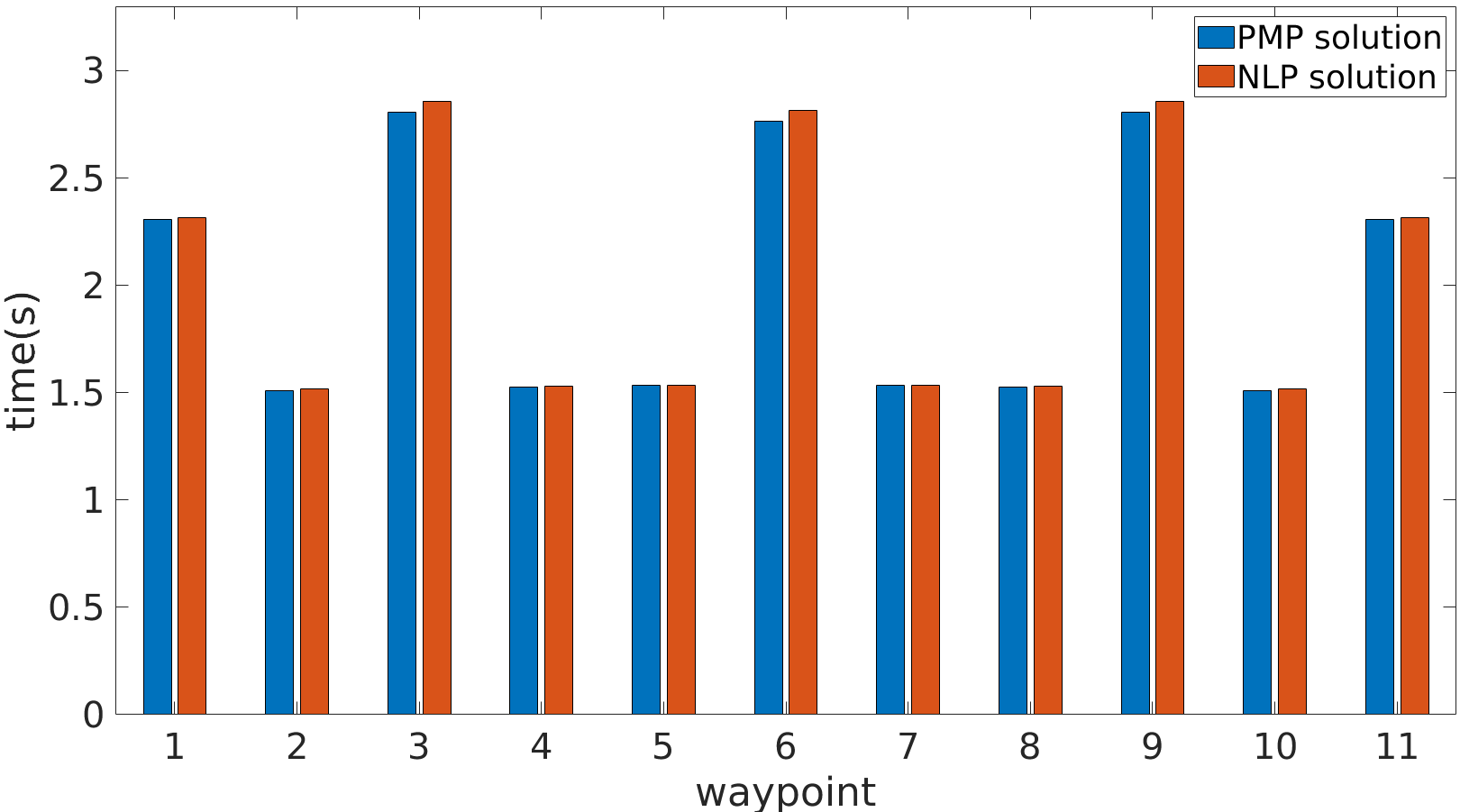}
    \caption{Comparison of time taken between consecutive waypoints using discretized and continuous-time optimal solution,i.e. ${t(w_{i}\rightarrow w_{i+1})}$ vs ${w_{i}}$}
    \label{fig:time_comp}
\end{figure}

\begin{figure}[!h]
    \centering
    \includegraphics[width=3.14in]{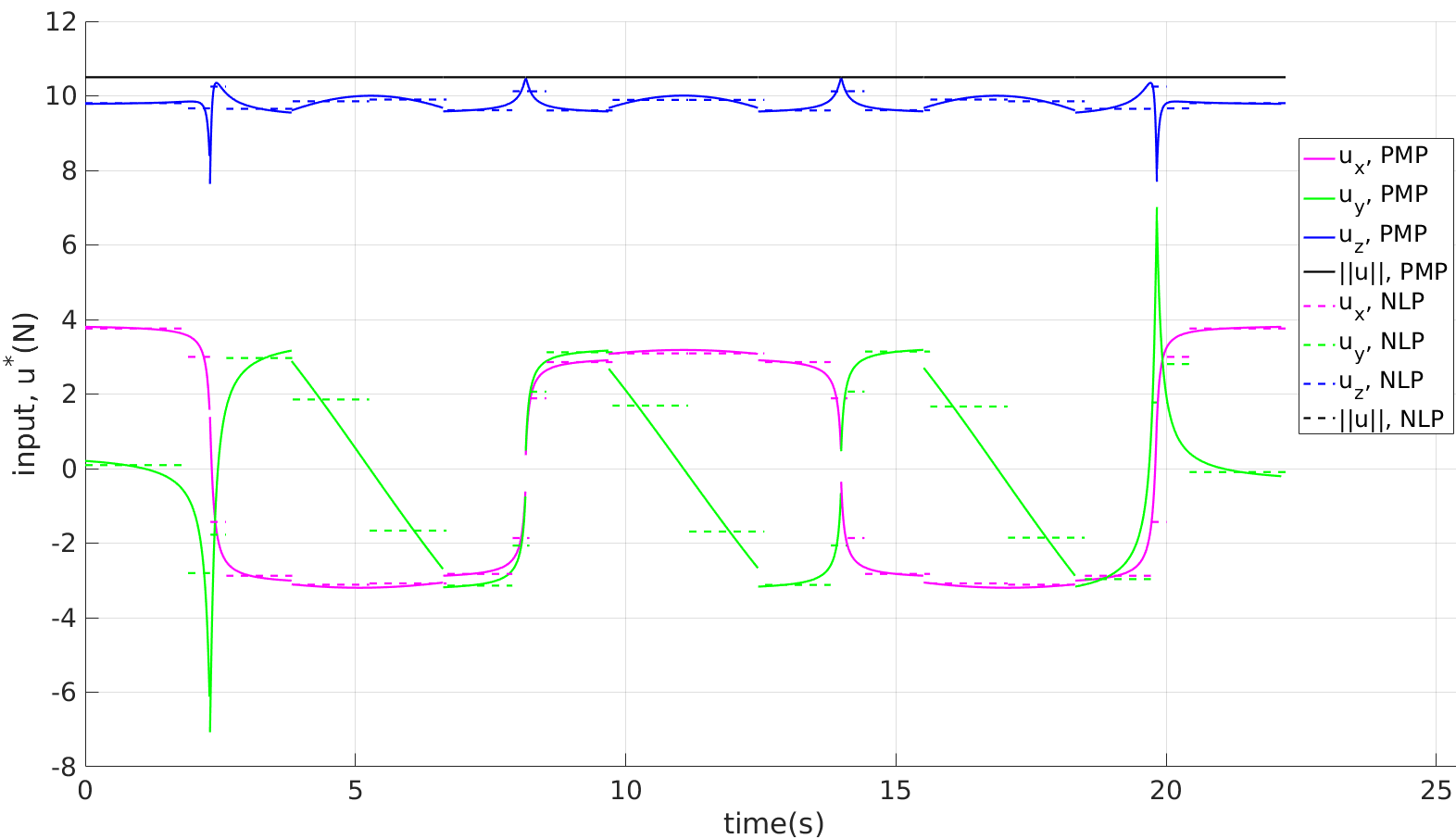}
    \caption{Optimal input functions obtained from direct interpolation and PMP approaches using 1 switching point}
    \label{fig:inp_comp}
\end{figure}

Figures \ref{fig:traj_comp} and \ref{fig:vel_comp} show a comparison of the trajectory obtained from direct interpolation v/s the trajectory obtained from solving the optimal control problem. It can be seen that the trajectory interpolations are not very different. Direct interpolation is easy to perform, incurs small computation cost, and gives a solution which is very close to the optimal one. Hence a discretized minimum-time control strategy produces a trajectory that is close to the continuous minimum-time optimal solution. The time taken to go from a waypoint $w_{i}\in W(A)$ to the next, $w_{i+1}\in W(A)$, is shown in in Figure \ref{fig:time_comp}. We observe that the times obtained from the NLP solution is not far from the solution obtained from the PMP-based approach. While the time saved by using the PMP approach is low, our method generates a much smoother trajectory with a continuous input through all waypoints as shown in Figures \ref{fig:inp_comp} and \ref{fig:inp_comp_4npts}. This makes it easier to implement our method on real systems. Works such as \cite{foehn2021time} propose using a time-optimal solution for the particle steering problem as an initial guess for the time-optimal steering of a UAV with it's rotary dynamics included. The approach here provides a continuous-time optimal input which potentially leads to better solutions after refinement for a UAV's underactuated dynamics.

The graph of the Hamiltonian v/s time shown in Figure \ref{fig:vel_comp} is a horizontal line between two consecutive waypoints but is not constant throughout the trajectory. This means that better waypoint velocities can exist that ensure the Hamiltonian remains constant across all waypoints, which is a requirement from PMP. That would make it the truly time-optimal solution. Refining the waypoint velocity guesses is a possible direction for future work.

\subsection{Influence of switching points}

\begin{figure}[ht]
    \centering
    \includegraphics[width=3.1in]{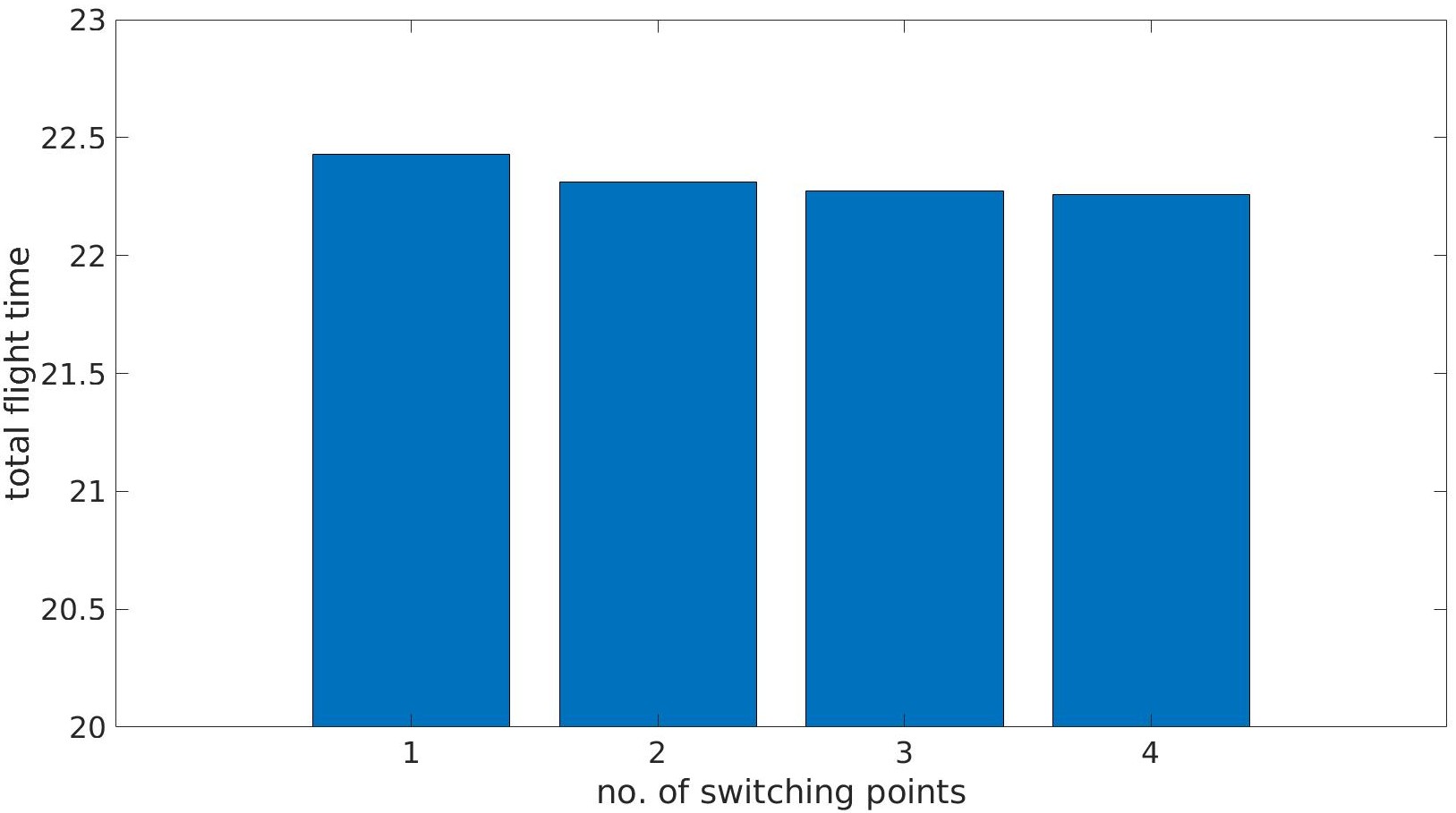}
    \caption{Using more switching points doesn't significantly impact the total flight time}
    \label{fig:npts_vs_time}
\end{figure}

\begin{figure}[ht]
    \centering
    \includegraphics[width=3.3in]{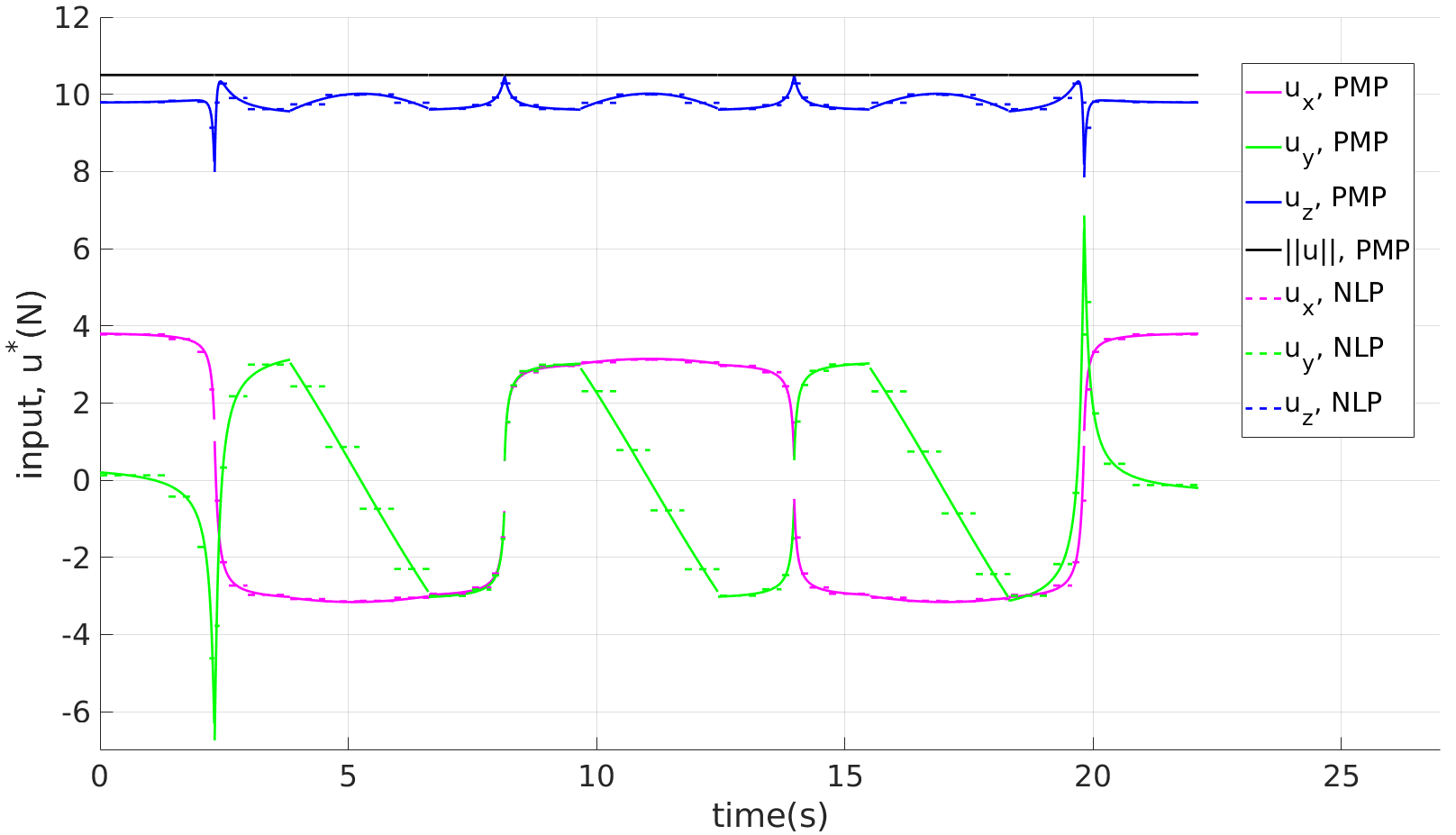}
    \caption{Optimal input functions obtained from direct interpolation and PMP approaches using 4 switching points}
    \label{fig:inp_comp_4npts}
\end{figure}

In our formulation, the input function is discretized by using switching points. Only 1 intermediate switching point was used but it is possible to have two or more switching points. The result of using more than one switching point is shown in Figure  \ref{fig:npts_vs_time}. Using multiple switching points doesn't greatly affect the final objective function. The final time difference between using one and multiple switching points is not significant. The inputs inferred from those velocities are nearly identical and appear slightly smoother near the waypoints. This can be seen when comparing Figure \ref{fig:inp_comp} and Figure \ref{fig:inp_comp_4npts}. Clearly, using more switching points increases the computation time of the solution to the NLP. For a large number of switching points, some solvers had difficulty converging to a solution. To reduce computation time, it is better to use one switching point. If the requirement is to have a smoother time-optimal input, then multiple switching points may be used.

\subsection{Comparison with alternative trajectory generation methods}
\begin{figure}[ht]
    \centering
    \includegraphics[width=3in, trim={0 0cm 0 0cm},clip]{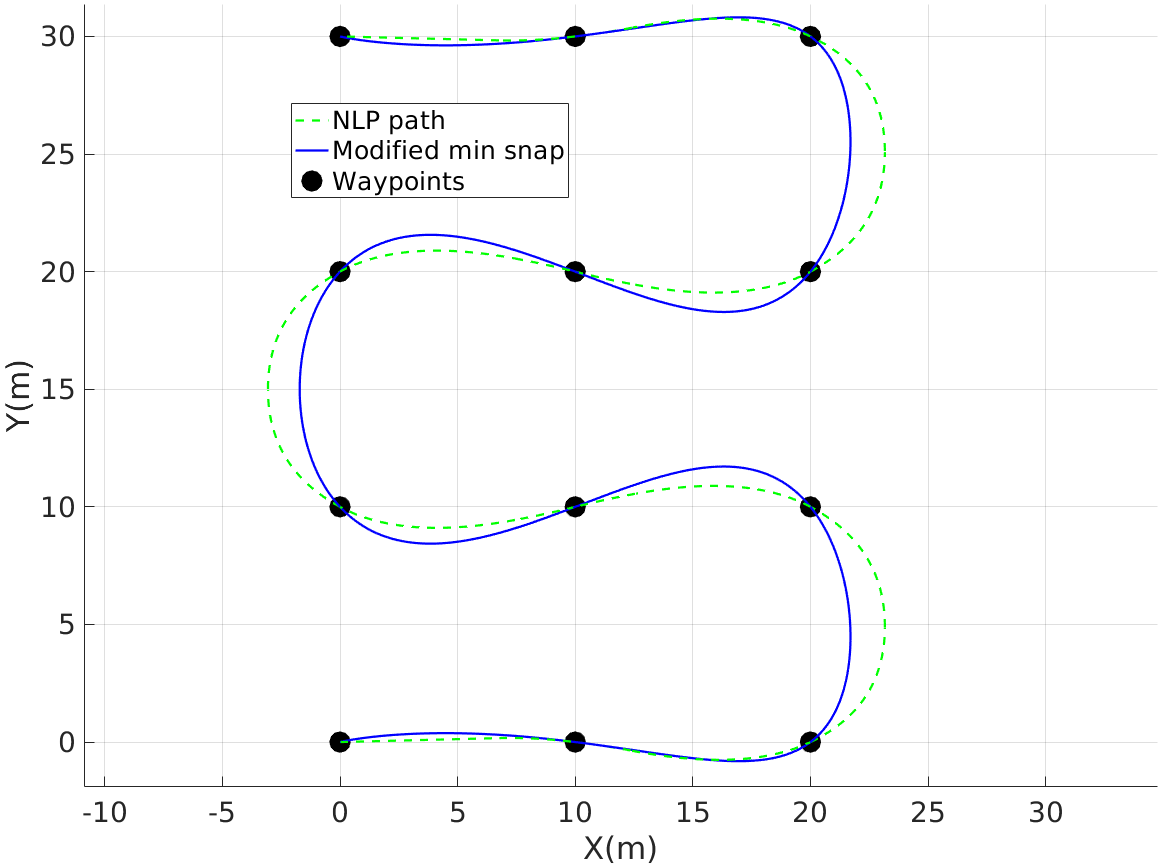}
    \caption{Comparison of minimum snap trajectory (solid) vs our trajectory generator (dotted)}
    \label{fig:traj_min_snap_comp}
\end{figure}

\begin{figure}[ht]
    \centering
    \includegraphics[width=3.14in, trim={0 0cm 0 0cm},clip]{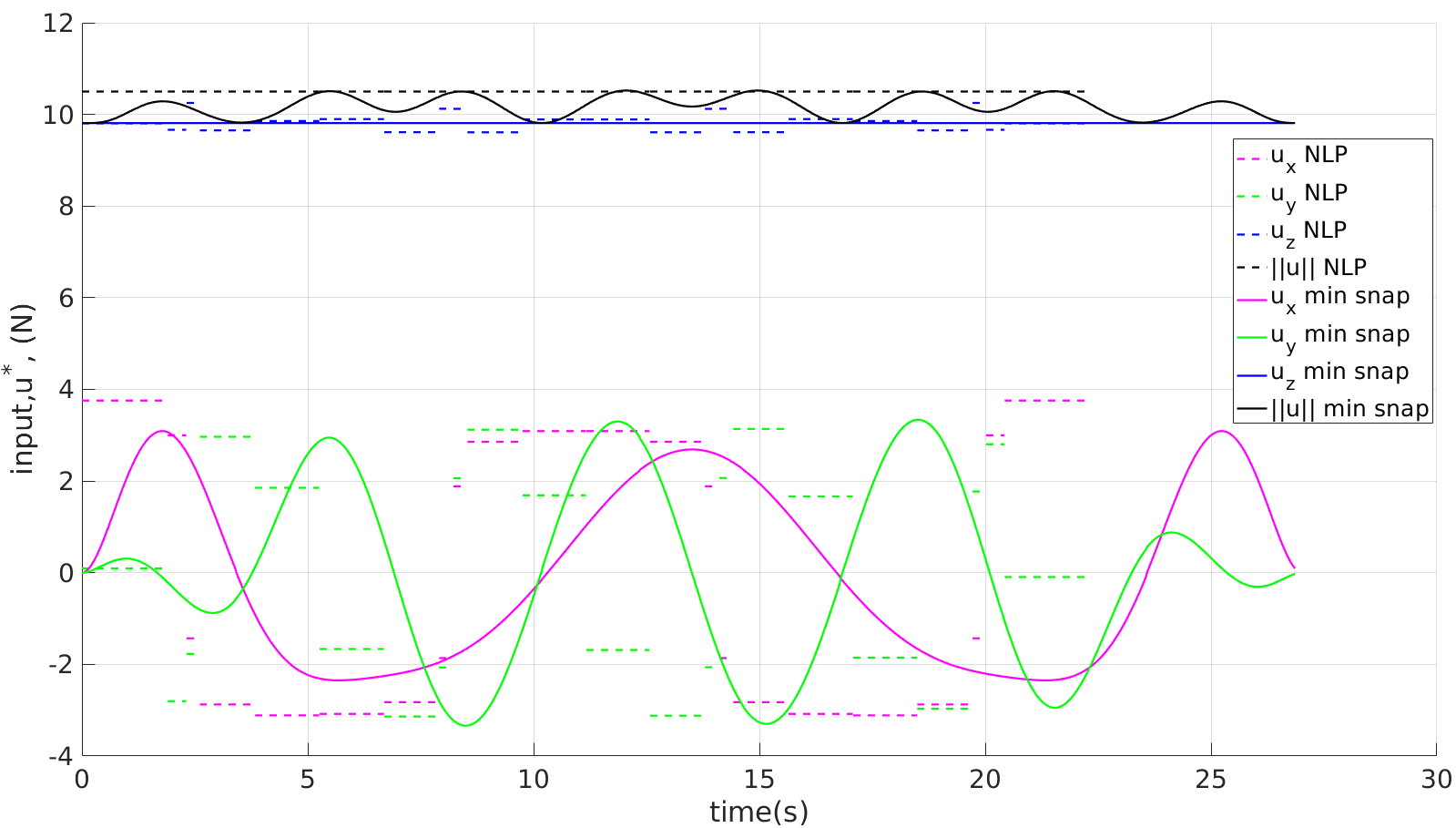}
    \caption{Comparison of modified min. snap trajectory (solid) vs the NLP trajectory generator (dotted)}
    \label{fig:inp_min_snap_comp}
\end{figure}

For a comparison of our algorithm to existing methods, we use the trajectory generated from the minimum snap problem described in \cite{mellinger2011minimum, ryou2020multi}. Most conventional UAV planning methods use a polynomial trajectory that minimizes jerk or snap. This is done to make the derivatives smooth and minimize body torques, but such polynomial planners do not ensure that the drone uses maximum actuator capacity/acceleration at all trajectory points, as seen in Figures \ref{fig:traj_min_snap_comp} and \ref{fig:inp_min_snap_comp}. Our approach produces a continuous input function while ensuring that the boundary of the input value set is achieved at all points in the trajectory, as seen in Figure \ref{fig:inp_min_snap_comp}. As a result, the total time taken by the minimum snap planner to complete the trajectory is 20\% slower than our method.

% \begin{figure}[!h]
%     \centering
%     \includegraphics[width=3.35in, trim={0 0cm 0 0cm},clip]{./figures/time_comp_NLP_snap.png}
%     \caption{Comparison of modified min. snap trajectory (blue) vs our trajectory generator (green)}
%     \label{fig:time_min_snap_comp}
% \end{figure}

\section{Conclusion}
\label{sec:conclusion}
In this work we presented a time-optimal surveying planner that generates a continuous trajectory which uses maximum thrust at all points and minimizes the aggregate flight time. Solution optimality is verified by comparing the generated trajectory to the solution obtained by solving the time-optimal control problem between two waypoints. 
We did this by discretizing and formulating a NLP problem for time-optimal surveying. We then use that to generate position and velocity references that can be directly interpolated to generate a continuous trajectory. We show that a direct interpolation is not far off from the optimal control input trajectory obtained by solving for the PMP solution.
Next we study the effect of using multiple switching points and how much improvement can be obtained from increasing them. No significant time savings were observed on adding more intermediate switching points. However, the continuous-time optimal input that was inferred using multiple switching points was slightly smoother.

While this work focuses on aggregate flight time, total computation time is another aspect that can be optimized for reduction in time. This can be addressed in a future work. Possible approaches include the use of a better initial guess(convexifying the NLP and using that solution) or using data-driven approaches, \cite{de2019real}, that learn from the planner described in this work. 

\section{Acknowledgement}
This work was funded by the Rig Automation and Performance Improvement in Drilling (RAPID) Consortium of University of Texas at Austin. 

\printbibliography

@article{loianno2018special,
  title={Special issue on high-speed vision-based autonomous navigation of uavs},
  author={Loianno, Giuseppe and Scaramuzza, Davide and Kumar, Vijay},
  journal={Journal of Field Robotics},
  volume={1},
  number={1},
  pages={1--3},
  year={2018},
  publisher={Wiley-Blackwell Publishing, Inc.}
}

@article{foehn2021time,
  title={Time-optimal planning for quadrotor waypoint flight},
  author={Foehn, Philipp and Romero, Angel and Scaramuzza, Davide},
  journal={Science Robotics},
  volume={6},
  number={56},
  year={2021},
  publisher={American Association for the Advancement of Science}
}

@article{bakolas2013optimal,
  title={Optimal synthesis of the Zermelo--Markov--Dubins problem in a constant drift field},
  author={Bakolas, Efstathios and Tsiotras, Panagiotis},
  journal={Journal of Optimization Theory and Applications},
  volume={156},
  number={2},
  pages={469--492},
  year={2013},
  publisher={Springer}
}

@Article{coverage2019,
AUTHOR = {Cabreira, Tauã M. and Brisolara, Lisane B. and Ferreira Jr., Paulo R.},
TITLE = {Survey on Coverage Path Planning with Unmanned Aerial Vehicles},
JOURNAL = {Drones},
VOLUME = {3},
YEAR = {2019},
NUMBER = {1},
ARTICLE-NUMBER = {4}
}

@article{ryou2020multi,
  title={Multi-fidelity black-box optimization for time-optimal quadrotor maneuvers},
  author={Ryou, Gilhyun and Tal, Ezra and Karaman, Sertac},
  journal={arXiv preprint arXiv:2006.02513},
  year={2020}
}

@inproceedings{mellinger2011minimum,
  title={Minimum snap trajectory generation and control for quadrotors},
  author={Mellinger, Daniel and Kumar, Vijay},
  booktitle={2011 ICRA},
  pages={2520--2525},
  year={2011},
  organization={IEEE}
}

@article{lupashin2011adaptive,
title = {Adaptive Open-Loop Aerobatic Maneuvers for Quadrocopters},
journal = {IFAC Proceedings Volumes},
volume = {44},
number = {1},
year = {2011},
note = {18th IFAC World Congress},
author = {Sergei Lupashin and Raffaello D'Andrea},
}

@inproceedings{de2019real,
title={Real-time minimum snap trajectory generation for quadcopters: Algorithm speed-up through machine learning},
author={de Almeida, Marcelino M and Moghe, Rahul and Akella, Maruthi},
booktitle={2019 ICRA},
year={2019}
}

@article{lai2006time,
  title={Time-optimal control of a hovering quad-rotor helicopter},
  author={Lai, Li-Chun and Yang, Chi-Ching and Wu, Chia-Ju},
  journal={Journal of Intelligent and Robotic Systems},
  volume={45},
  number={2},
  pages={115--135},
  year={2006},
  publisher={Springer}
}

@inproceedings{van2013time,
  title={Time-optimal quadrotor flight},
  author={Van Loock, Wannes and Pipeleers, Goele and Swevers, Jan},
  booktitle={2013 European Control Conference (ECC)},
  year={2013},
  organization={IEEE}
}

@article{bakolas2016time,
  title={Time-optimal control of a self-propelled particle in a spatiotemporal flow field},
  author={Bakolas, Efstathios and Marchidan, Andrei},
  journal={International Journal of Control},
  volume={89},
  number={3},
  pages={623--634},
  year={2016},
  publisher={Taylor \& Francis}
}

@inproceedings{venkatraman2006optimal,
  title={Optimal planar turns under acceleration constraints},
  author={Venkatraman, Aneesh and Bhat, Sanjay P},
  booktitle={Proceedings of the 45th IEEE Conference on Decision and Control},
  pages={235--240},
  year={2006}
}

@article{akulenko2002timesphere,
  title={Time-optimal steering of a point mass onto the surface of a sphere at zero velocity},
  author={Akulenko, LD and Shmatkov, AM},
  journal={Journal of Applied Mathematics and Mechanics},
  volume={66},
  number={1},
  pages={9--21},
  year={2002},
  publisher={Elsevier}
}

@techreport{lee1967foundations,
  title={Foundations of Optimal Control Theory},
  author={Lee, Ernest Bruce and Markus, Lawrence},
  year={1967},
  institution={Minnesota Univ Minneapolis Center For Control Sciences}
}

@article{akulenko2007time,
  title={Time-optimal steering of a point mass to a specified position with the required velocity},
  author={Akulenko, LD and Koshelev, AP},
  journal={Journal of Applied Mathematics and Mechanics},
  volume={71},
  number={2},
  pages={200--207},
  year={2007},
  publisher={Elsevier}
}

@article{ZHOU202056,
title = {Survey on path and view planning for UAVs},
journal = {Virtual Reality and Intelligent Hardware},
volume = {2},
number = {1},
pages = {56-69},
year = {2020},
issn = {2096-5796},
author = {Xiaohui Zhou and Zimu Yi and Yilin Liu and Kai Huang and Hui Huang}
}

@inproceedings{du2018boundary,
  title={Boundary Control of a Quadrotor UAV with a Payload Connected by a Flexible Cable},
  author={Du, Zhong-Hui and Wu, Huai-Ning and Feng, Shuang},
  booktitle={2018 37th Chinese Control Conference (CCC)},
  pages={1151--1156},
  year={2018},
  organization={IEEE}
}

@inproceedings{traj:benchmark,
  title={Quadrocopter performance benchmarking using optimal control},
  author={Ritz, Robin and Hehn, Markus and Lupashin, Sergei and D'Andrea, Raffaello},
  booktitle={2011 IEEE/RSJ International Conference on Intelligent Robots and Systems},
  year={2011}
}

@book{athans2013optimal,
  title={Optimal control: an introduction to the theory and its applications},
  author={Athans, Michael and Falb, Peter L},
  year={2013},
  publisher={Courier Corporation}
}

@article{IPOPT,
  title={On the implementation of an interior-point filter line-search algorithm for large-scale nonlinear programming},
  author={W{\"a}chter, Andreas and Biegler, Lorenz T},
  journal={Mathematical programming},
  volume={106},
  number={1},
  pages={25--57},
  year={2006},
  publisher={Springer}
}

\end{document}